\documentclass[10pt,twocolumn,letterpaper]{article}

\usepackage[pagenumbers]{cvpr}      
\definecolor{cvprblue}{rgb}{0.21,0.49,0.74}
\usepackage[pagebackref,breaklinks,colorlinks,allcolors=cvprblue]{hyperref}
\usepackage{multirow}
\usepackage{pifont}

\def\paperID{9864} 
\def\confName{CVPR}
\def\confYear{2026}

\title{DiffusionFF: A Diffusion-based Framework for Joint Face Forgery Detection and Fine-Grained Artifact Localization}

\author{
    Siran Peng\textsuperscript{\rm 1,\,2}\thanks{Equal Contribution.} \quad
    Haoyuan Zhang\textsuperscript{\rm 2,\,1}\footnotemark[1] \quad
    Li Gao\textsuperscript{\rm 3} \quad
    Tianshuo Zhang\textsuperscript{\rm 2,\,1} \\
    Xiangyu Zhu\textsuperscript{\rm 1,\,2} \quad
    Bao Li\textsuperscript{\rm 1,\,2} \quad
    Weisong Zhao\textsuperscript{\rm 4,\,5} \quad
    Zhen Lei\textsuperscript{\rm 1,\,2,\,6,\,7}\thanks{Corresponding author.} \\ 
    \small \rm 
    \textsuperscript{1}MAIS, CASIA \quad
    \textsuperscript{2}SAI, UCAS \quad
    \textsuperscript{3}CMFT \quad
    \textsuperscript{4}IIE, CAS \quad
    \textsuperscript{5}SCS, UCAS \quad
    \textsuperscript{6}CAIR, HKISI, CAS \quad
    \textsuperscript{7}SCSE, FIE, M.U.S.T \\
    {\tt\small \{pengsiran2023, zhanghaoyuan2023, zhen.lei\}@ia.ac.cn}
}

\begin{document}
\maketitle
\begin{abstract}
The rapid evolution of deepfake technologies demands robust and reliable face forgery detection algorithms. While determining whether an image has been manipulated remains essential, the ability to precisely localize forgery clues is also important for enhancing model explainability and building user trust. To address this dual challenge, we introduce DiffusionFF, a diffusion-based framework that simultaneously performs face forgery detection and fine-grained artifact localization. Our key idea is to establish a novel encoder–decoder architecture: a pretrained forgery detector serves as a powerful ``artifact encoder", and a denoising diffusion model is repurposed as an ``artifact decoder". Conditioned on multi-scale forgery-related features extracted by the encoder, the decoder progressively synthesizes a detailed artifact localization map. We then fuse this fine-grained localization map with high-level semantic features from the forgery detector, leading to substantial improvements in detection capability. Extensive experiments show that DiffusionFF achieves state-of-the-art (SOTA) performance across multiple benchmarks, underscoring its superior effectiveness and explainability.
\end{abstract}
\section{Introduction}
The rapid advancement of deepfake technologies \cite{Nirkin_2019_ICCV,9320155,Hsu_2022_CVPR,Shiohara_2023_ICCV} has enabled the creation of forged facial content with highly convincing perceptual quality, posing serious threats to information integrity and public trust. This escalating concern underscores the urgent demand for robust and reliable face forgery detection algorithms. Traditionally, research in this area focuses on the binary classification task of determining whether an image has been manipulated. However, there is a growing recognition of the need to precisely localize forgery clues. By pinpointing the manipulation traces within an image, artifact localization offers clear, human-interpretable evidence that supports the model's decisions, thereby improving explainability and fostering greater user trust. To simultaneously achieve both objectives, recent studies are increasingly exploring unified frameworks that integrate detection and artifact localization.

\begin{figure}[t]
	\begin{center}
		\begin{minipage}{1\linewidth}
			{\includegraphics[width=1\linewidth]{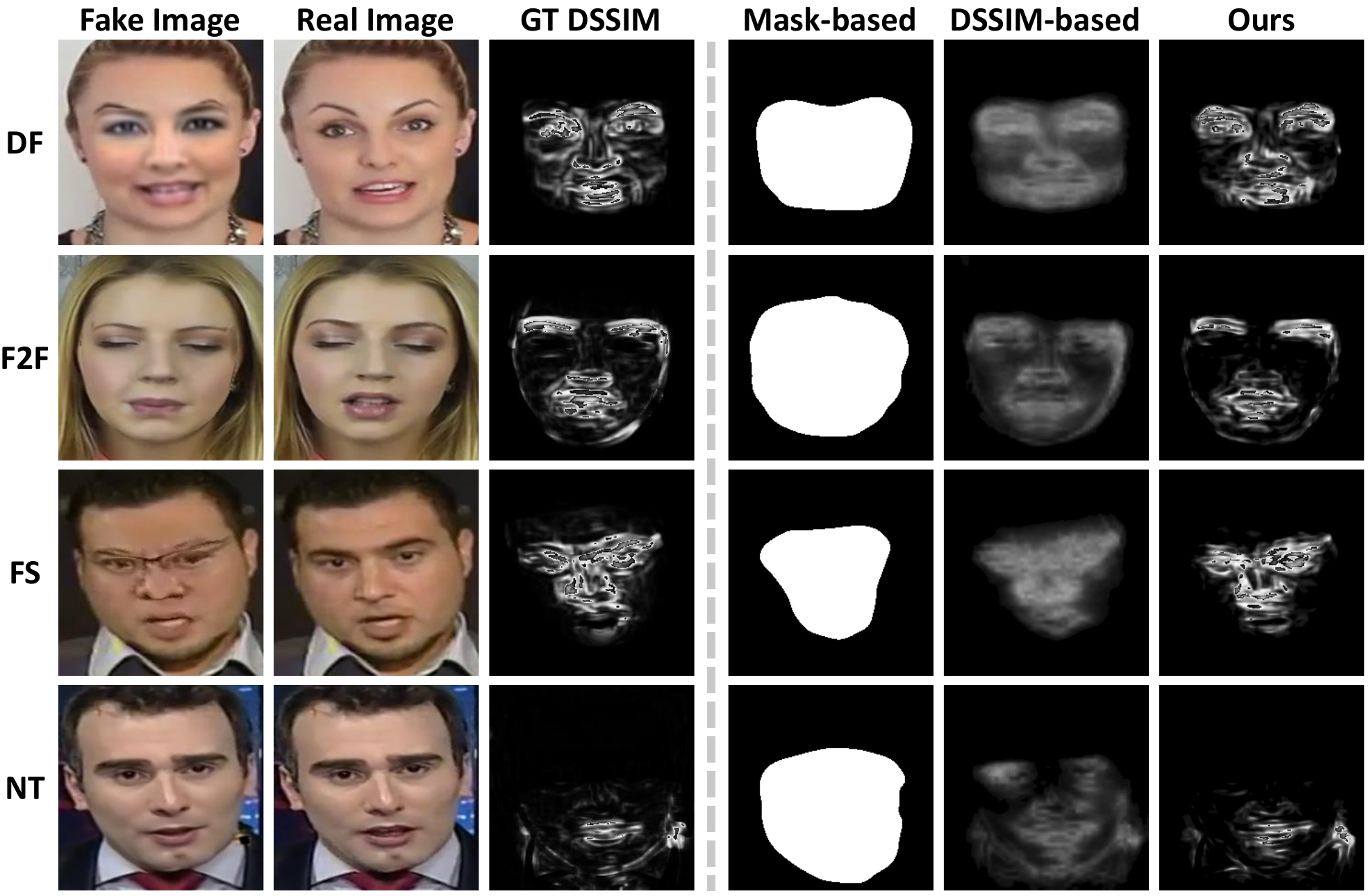}}
			\centering
		\end{minipage}
	\end{center}
    \caption{Visual comparison of DiffusionFF with mask-based and existing DSSIM-based artifact localization methods across four manipulation types (DF, F2F, FS, and NT) from the FaceForensics++ (FF++) dataset \cite{Rossler_2019_ICCV}. The Ground-Truth (GT) DSSIM map is generated by comparing each fake image with its corresponding real image using the algorithm detailed in Section~\ref{gt_ssim}. \label{hp}}
\end{figure}

However, the precise localization of fine-grained forgery clues remains a significant challenge. Conventional mask-based methods \cite{kong2022detect, zhao2023hybrid, 10.24963/ijcai.2024/648} are inherently coarse, highlighting only broad regions of manipulation. To overcome this limitation, several studies have explored using Structural Dissimilarity (DSSIM) maps \cite{ssim} to guide localization map generation. By directly comparing pixel-level differences between aligned real and fake image pairs, DSSIM maps effectively capture subtle forgery artifacts. Despite this potential, existing DSSIM-based approaches \cite{chen2021local, wang2022lisiam} are constrained by their reliance on direct regression frameworks, which tend to smooth over subtle manipulation traces, resulting in blurry and less informative localization maps. A visual comparison of these methods is presented in Figure~\ref{hp}.

\begin{figure}[t]
	\begin{center}
		\begin{minipage}{1\linewidth}
			{\includegraphics[width=0.93\linewidth]{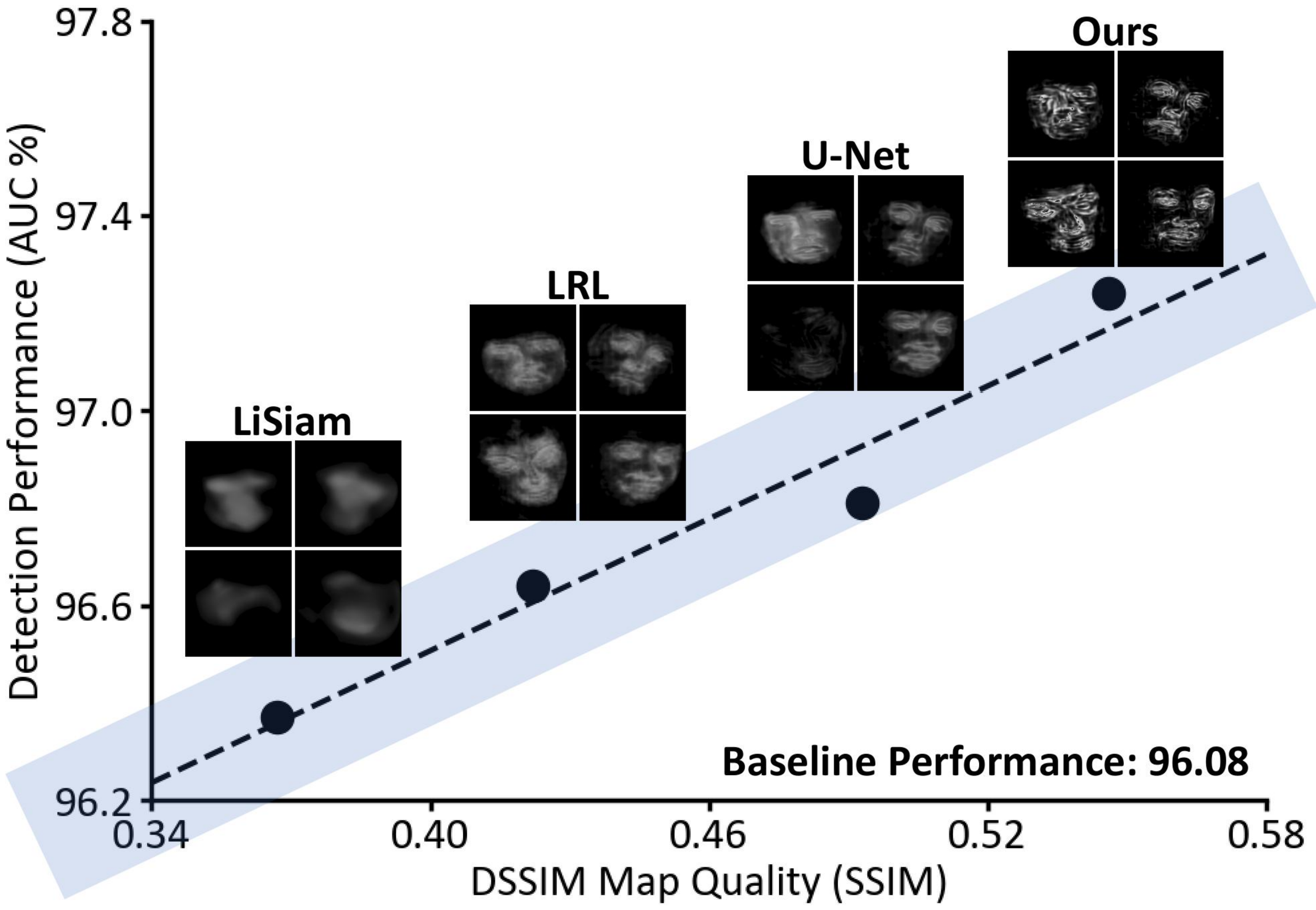}}
			\centering
		\end{minipage}
	\end{center}
	\caption{Correlation between the quality of the estimated DSSIM maps and detection performance. We fuse the DSSIM maps estimated by LiSiam \cite{wang2022lisiam}, LRL \cite{chen2021local}, U-Net \cite{ronneberger2015u}, and our DiffusionFF with high-level semantic features from a shared forgery detector to obtain classification results. Integrating estimated DSSIM maps into the detection network consistently improves performance, with higher-quality maps leading to greater performance gains.
    \label{relation}}
\end{figure}

In \cite{wang2022lisiam}, the estimated DSSIM maps are reintegrated into the network to enhance detection capability. This strategy serves as a key motivation for our work. As shown in Figure~\ref{relation}, our analysis reveals a strong correlation between the quality of these maps and overall detection performance. Specifically, incorporating estimated DSSIM maps into the detection network consistently improves performance, with higher-quality maps leading to greater performance gains.

Considering these factors, we propose DiffusionFF, a novel framework designed to simultaneously tackle the dual challenges of face forgery detection and fine-grained artifact localization. The core innovation of DiffusionFF lies in its rational and intuitive encoder–decoder architecture, where a pretrained forgery detector functions as a powerful ``artifact encoder", and a denoising diffusion model \cite{ddpm} is repurposed as an ``artifact decoder". Conditioned on multi-scale forgery-relevant features extracted by the encoder, the decoder progressively estimates a detailed DSSIM map, which precisely localizes fine-grained forgery clues. This estimated DSSIM map is subsequently fused with high-level semantic features from the forgery detector, leading to substantial improvements in detection performance. In conclusion, the contributions of this paper are as follows:

\begin{itemize}
\item We establish a novel encoder-decoder architecture that leverages a pre-trained forgery detector as an ``artifact encoder" and repurposes a denoising diffusion model as an ``artifact decoder". Conditioned on multi-scale forgery-related features from the encoder, the decoder progressively synthesizes a fine-grained artifact localization map.
\item We integrate the generated localization map with high-level semantic features from the forgery detector, resulting in significant improvements in detection capability.
\item DiffusionFF achieves state-of-the-art (SOTA) detection performance and surpasses existing artifact localization methods in revealing manipulation traces, demonstrating its superior effectiveness, reliability, and explainability.
\end{itemize}
\section{Related Works \& Motivation}
\subsection{Face Forgery Detection}
\subsubsection{Detection-only Methods} 
Detection-only methods for face forgery detection primarily focus on achieving high accuracy, typically only providing a binary classification outcome. Existing studies, based on their underlying strategies, can be grouped into four main categories \cite{pei2024deepfake}: spatial-domain, time-domain, frequency-domain, and data-driven approaches. Spatial-domain methods detect forgeries by analyzing image-level artifacts such as color mismatches \cite{8803740}, unnatural saturation \cite{8803661}, and visual anomalies \cite{Zhao_2021_CVPR,9694644,cui2024forensics}. Time-domain techniques model temporal dynamics across video frames to capture inter-frame inconsistencies \cite{10054130,zhang2025learning}. Frequency-domain methods apply algorithms like the Fast Fourier Transform (FFT) \cite{tan2024frequency}, Discrete Cosine Transform (DCT) \cite{qian2020thinking}, and Discrete Wavelet Transform (DWT) \cite{9447758,10.1145/3503161.3547832,peng2025wmamba} to expose subtle manipulation traces that are often imperceptible in the spatial or time domains. Lastly, data-driven approaches boost detection performance by leveraging advanced network architectures and optimized training strategies \cite{Zhao_2021_ICCV,Yan_2023_ICCV,Huang_2023_CVPR,Yan_2024_CVPR}. 

\subsubsection{Joint Detection and Localization Methods}
In addition to detection-only approaches, some studies have developed unified frameworks capable of simultaneously performing face forgery detection and artifact localization. These methods can be broadly classified into two types: mask-based and DSSIM-based approaches. Mask-based methods treat artifact localization as a segmentation task, generating binary masks that indicate forged regions. These methods primarily exploit semantic features, analyze noise patterns, or leverage meta-learning techniques to improve segmentation precision \cite{kong2022detect, zhao2023hybrid, 10.24963/ijcai.2024/648}. In contrast, DSSIM-based approaches aim to produce fine-grained, pixel-level localization maps that reveal subtle traces of manipulation. To enhance localization quality, these approaches have explored attention mechanisms \cite{chen2021local} or Siamese network architectures \cite{wang2022lisiam}. Unlike coarse binary masks, DSSIM maps offer a more precise and detailed representation of forgery artifacts. However, existing DSSIM-based methods rely on direct regression frameworks, leading to blurry and overly smoothed outputs. This limitation prevents these methods from fully harnessing the rich potential of DSSIM maps.

\subsection{Diffusion Models in Forgery Detection}
In recent years, diffusion models \cite{ddpm} have emerged as a powerful class of generative tools, which iteratively transform noise into high-fidelity images by learning the underlying data distribution through a denoising process. In image forgery detection, DiffForensics \cite{yu2024diffforensics} leverages diffusion models as robust image priors to detect manipulations and create binary masks highlighting forged regions. In the specific domain of face forgery detection, DiffusionFake \cite{chen2024diffusionfake} employs a Latent Diffusion Model (LDM) \cite{rombach2022high} to reconstruct both the source and target identities from a single deepfake image. However, despite these advancements, the potential of diffusion models to synthesize fine-grained artifact localization maps remains largely unexplored.

\subsection{Motivation}
Recent studies are increasingly exploring unified frameworks that jointly address face forgery detection and artifact localization. Conventional mask-based methods provide only coarse localization of forged regions. On the other hand, while DSSIM maps can capture fine-grained manipulation traces, existing DSSIM-based approaches often produce blurry outputs due to their reliance on direct regression frameworks. To overcome these limitations and fully exploit the potential of DSSIM maps, we establish a novel encoder-decoder architecture that leverages a pre-trained forgery detector as an ``artifact encoder" and reproposes a diffusion model as an ``artifact decoder". Conditioned on forgery-relevant features from the encoder, the decoder progressively estimates a detailed DSSIM map, enabling precise localization of fine-grained artifacts. Motivated by our analysis in Figure~\ref{relation}, we fuse the estimated DSSIM map with high-level semantic features from the detector, leading to substantial improvements in detection performance. 

\begin{figure*}[t]
	\begin{center}
		\begin{minipage}{1\linewidth}
			{\includegraphics[width=1\linewidth]{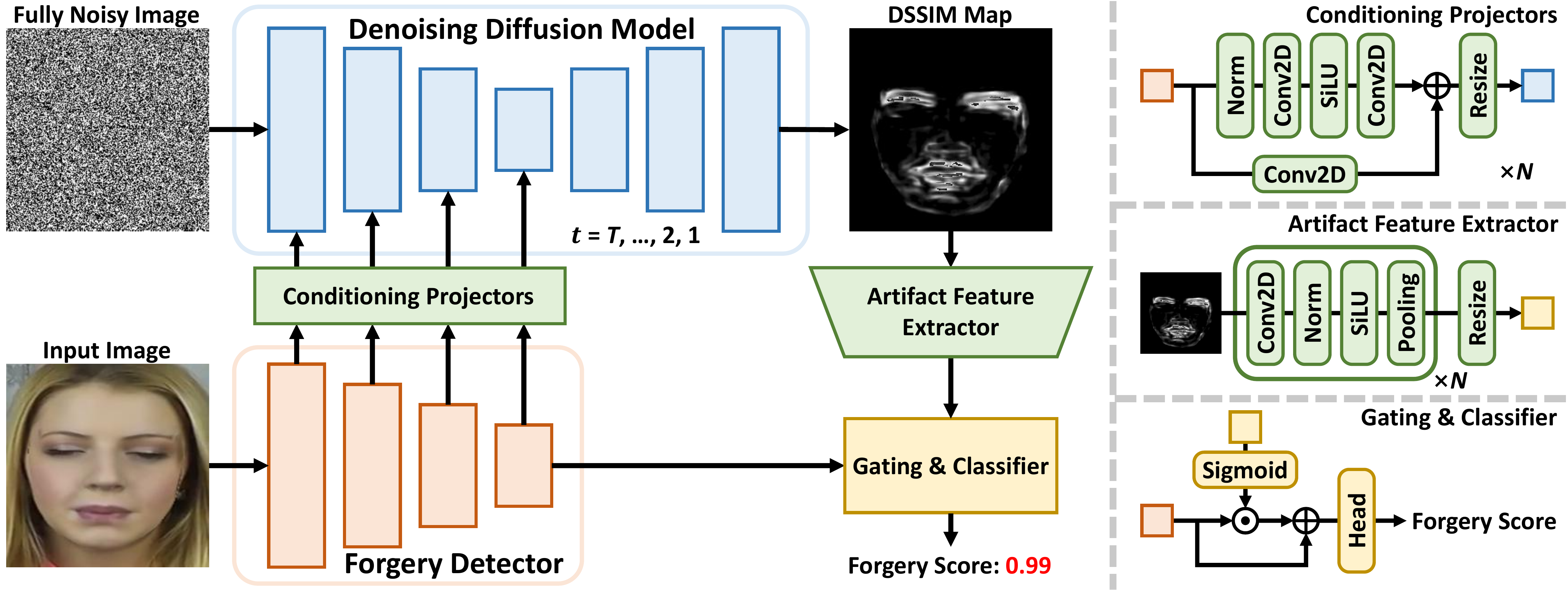}}
			\centering
		\end{minipage}
	\end{center}
	\caption{Overview of the DiffusionFF framework. Given an input facial image, DiffusionFF simultaneously predicts a forgery score and estimates a DSSIM map that precisely localizes fine-grained facial forgery clues. $N$ denotes the number of stages in the forgery detector. \label{pipeline}}
\end{figure*}
\section{Methodology}
\subsection{Preliminaries}
\subsubsection{GT DSSIM Map Formulation}\label{gt_ssim}
To generate Ground-Truth (GT) DSSIM maps for supervising the training of denoising diffusion models and for evaluation purposes, we first extract frames and crop faces from both the original facial video and its manipulated counterparts, ensuring consistent settings across different video types. This process yields a set of fully aligned image pairs. For each image pair, we compute the DSSIM map by sliding a local square window across the two images and calculating the DSSIM value at each pixel location $(i, j)$. The computation can be mathematically expressed as follows:
\begin{equation}\label{eq1}
\begin{aligned}
\text{DSSIM}(i, j) &= 1 - \text{SSIM}(x, y), \\
\text{SSIM}(x, y) &= \frac{(2\mu_x\mu_y+C_1)(2\sigma_{xy}+C_2)}{(\mu^2_x+\mu^2_y+C_1)(\sigma^2_x+\sigma^2_y+C_2)}.
\end{aligned}
\end{equation}
Here, $\text{DSSIM}(i, j)$ is the DSSIM value at the pixel position $(i, j)$. The variables $x$ and $y$ represent the local square windows centered at $(i, j)$ in the original and manipulated images. $\mu_x$ and $\mu_y$ are the mean intensities of windows $x$ and $y$, while $\sigma_x$ and $\sigma_y$ denote the variances of the corresponding windows. $\sigma_{xy}$ represents the covariance between the two windows. The constants $C_1$ and $C_2$ are small values added to ensure numerical stability. Notably, DSSIM maps are computed only for fake images, while real images are assigned pure black maps to represent zero dissimilarity.

\subsubsection{Diffusion Model Training} 
We train our generative model following the Denoising Diffusion Probabilistic Model (DDPM) framework \cite{ddpm}, which involves two primary processes: a forward diffusion process and a reverse denoising process. The forward process is a fixed Markov chain that progressively adds Gaussian noise to a clean image $x_0$ over $T$ timesteps. This process has a key property that a noisy image $x_t$ at any timestep $t$ can be directly sampled from the original image $x_0$ as follows:
\begin{equation}\label{eq2}
x_t = \sqrt{\bar{\alpha}_t} x_0 + \sqrt{1 - \bar{\alpha}_t} \epsilon,
\end{equation}
where $\epsilon \sim \mathcal{N}(0, \mathbf{I})$ is a random noise sample. The noise level is dictated by a predefined variance schedule $\{\beta_t\}_{t=1}^T$. From this, we define $\alpha_t = 1 - \beta_t$ and the cumulative product $\bar{\alpha}_t = \prod_{i=1}^t \alpha_i$. The reverse process aims to train a denoising network $\epsilon_\theta(x_t, t)$ that predicts the noise $\epsilon$ added to the noisy image $x_t$. The training objective is to minimize the Mean Squared Error (MSE) between the predicted noise and the true noise, which can be expressed as follows:
\begin{equation}\label{eq3}
    \mathcal{L}_{\text{MSE}} = \mathbb{E}_{x_0, \epsilon, t}\left[\left\|\epsilon - \epsilon_\theta(x_t, t)\right\|_2^2\right].
\end{equation}


\subsubsection{Diffusion Model Inference} 
During inference, the trained generative model reverses the forward diffusion process to reconstruct clean images from noisy inputs. Starting from a fully noisy image $x_T$, the model iteratively applies the reverse denoising process to refine it, eventually recovering the clean image $x_0$. At each timestep $t$, the denoising network $\epsilon_\theta(x_t, t)$ predicts the noise $\epsilon$ added to the image. The model then removes the predicted noise from the current noisy image $x_t$ to produce a less noisy version $x_{t-1}$, using the following update rule:
\begin{equation}\label{eq4}
\small
x_{t-1} = \frac{1}{\sqrt{\alpha_t}}\left(x_t-\frac{\beta_t}{\sqrt{1 - \bar{\alpha}_t}}\epsilon_\theta(x_t, t)\right) + \sqrt{\frac{1 - \bar{\alpha}_{t-1}}{1 - \bar{\alpha}_t}\beta_t}\cdot z,
\end{equation}
where $z\sim \mathcal{N}(0, \mathbf{I})$ is a random noise sample when $t>1$. This process is repeated for each $t$ from $T$ down to 1, gradually refining the image until a clean sample is generated.

\subsection{DiffusionFF}
\subsubsection{Overall Architecture}
The proposed DiffusionFF framework simultaneously performs face forgery detection and fine-grained artifact localization, as illustrated in Figure~\ref{pipeline}. Given an input facial image, a pretrained forgery detector first extracts multi-scale features that encode forgery-related information. These features are then injected into the diffusion model via conditioning projectors to guide the generative process. Starting from random noise, the diffusion model progressively estimates a detailed DSSIM map that precisely localizes fine-grained forgery clues. An artifact feature extractor subsequently extracts informative features from the estimated DSSIM map. Finally, these artifact-aware features are integrated with high-level features from the detector through a gating mechanism to produce the classification outcome.

\subsubsection{Conditional DSSIM Map Estimation via Diffusion} 
In previous studies, direct regression frameworks have been used to estimate DSSIM maps. However, such approaches often oversmooth subtle manipulation traces, resulting in blurry and less informative artifact localization. In contrast, diffusion models inherently support iterative refinement, allowing them to progressively enhance outputs and better capture fine-grained inconsistencies. Motivated by this advantage, we leverage a denoising diffusion model to estimate detailed DSSIM maps. To guide this generative process, we employ a pretrained forgery detector, which functions as a powerful ``artifact encoder", to extract multi-scale features from the input facial image. Conditioned on these forgery-relevant features, the diffusion model is thus repurposed as an ``artifact decoder", progressively synthesizing a fine-grained artifact localization map. Notably, we observe that jointly training the detector and diffusion model from scratch leads to training collapse, underscoring the importance of incorporating pre-learned forgery knowledge into the generation pipeline. To achieve this effective knowledge transfer, we design a set of conditioning projectors. These modules bridge the gap between the two models by aligning the spatial and channel dimensions of the detector's multi-scale features with the U-Net backbone \cite{ronneberger2015u} of the diffusion model. Once aligned, the forgery-related features are injected into the corresponding U-Net encoder stages, allowing the denoising diffusion model to fully exploit the detector's guidance and estimate detailed DSSIM maps.

\subsubsection{Feature Fusion and Classification} 
Figure~\ref{relation} demonstrates that integrating the estimated DSSIM maps into the detection network consistently improves performance. Therefore, we leverage the estimated high-fidelity DSSIM map to guide the final classification. Specifically, an artifact feature extractor first encodes the DSSIM map into artifact-aware features that capture subtle manipulation traces. These features are spatially and channel-wise aligned with the high-level semantic features from the final stage of the pretrained forgery detector. The two complementary feature sets are then fused via a gating mechanism, allowing the model to selectively emphasize manipulation-relevant regions. Finally, a classifier head produces the prediction. The overall process can be expressed as follows:
\begin{equation}\label{eq5}
\text{Score} = \mathcal{H}\bigl(\sigma\left(\mathcal{E}(M_{DSSIM})\right)\odot F_{det} + F_{det}\bigr),
\end{equation}
where $M_{DSSIM}$ is the estimated DSSIM map, and $F_{det}$ represents the detector's high-level features. $\mathcal{E}$ denotes the artifact feature extractor, $\sigma$ is the Sigmoid activation, $\odot$ represents the Hardamard product, and $\mathcal{H}$ is the classifier head.

\begin{table*}[t]	
	\centering\renewcommand\arraystretch{1.2}\setlength{\tabcolsep}{11.5pt}
	\belowrulesep=0pt\aboverulesep=0pt
        \caption{Cross-dataset face forgery detection performance on the CDF2, DFDC, DFDCP, and FFIW benchmarks. Methods marked with $\ast$ are reproduced, while results for others are directly cited from the original papers. For each dataset, the best result is highlighted in \textbf{bold} and the second-best is \underline{underlined}. Methods above the horizontal divider are detection-only, while those below are joint detection and localization approaches. Notably, the proposed DiffusionFF framework consistently achieves SOTA performance across all datasets. \label{cross-data}}
	\begin{tabular}{c|c|cc|cccc}
		\toprule
		\multirow{2}{*}{Method} & 
		\multirow{2}{*}{Venue} & 
		\multicolumn{2}{c|}{Localization} & 
		\multicolumn{4}{c}{Test Set AUC (\%) $\uparrow$}\\
		\cmidrule(lr){3-4}\cmidrule(lr){5-8}
		&\multicolumn{1}{c|}{} 
            &\multicolumn{1}{c}{Mask} 
            &\multicolumn{1}{c|}{DSSIM} 
		&\multicolumn{1}{c}{CDF2} 
		&\multicolumn{1}{c}{DFDC} 
		&\multicolumn{1}{c}{DFDCP}
		&\multicolumn{1}{c}{FFIW} \\
		\midrule
		 DCL \cite{sun2022dual} & AAAI 2022 & \ding{55} & \ding{55} & 82.30 & - & 76.71 & 71.14 \\
		 SBI \cite{Shiohara_2022_CVPR} & CVPR 2022 & \ding{55} & \ding{55} & 93.18 & 72.42 & 86.15 & {84.83} \\
		 F\textsuperscript{2}Trans \cite{10004978} & TIFS 2023 & \ding{55} & \ding{55} & 89.87 & - & 76.15 & - \\
		 SeeABLE \cite{Larue_2023_ICCV} & ICCV 2023 & \ding{55} & \ding{55} & 87.30 & 75.90 & 86.30 & - \\
		 LAA-Net \cite{Nguyen_2024_CVPR} & CVPR 2024 & \ding{55} & \ding{55} & 95.40 & - & 86.94 & - \\
		 RAE \cite{10.1007/978-3-031-72943-0_23} & ECCV 2024 & \ding{55} & \ding{55} &  {95.50} & {80.20} & 89.50 & - \\
		 FreqBlender \cite{zhou2024freqblender} & NeurIPS 2024 & \ding{55} & \ding{55}  & 94.59 & 74.59 & 87.56 & {86.14} \\
		 UDD \cite{fu2025exploring} & AAAI 2025 & \ding{55} & \ding{55} & 93.10 & 81.20 & 88.10 & - \\
         DFD-FCG \cite{Han_2025_CVPR} & CVPR 2025 & \ding{55} & \ding{55} & 95.00 & {81.81} & - & - \\
         Effort\textsuperscript{$\ast$} \cite{yan2024orthogonal} & ICML 2025 & \ding{55} & \ding{55} & \underline{95.73} & \underline{84.78} & 90.42 & \underline{88.53} \\ 
	\midrule
        LRL\textsuperscript{$\ast$} \cite{chen2021local} & AAAI 2021 & \ding{55} & \ding{51}  & 91.70 & 76.66 & 81.18 & 82.00 \\
        PCL+I2G \cite{Zhao_2021_ICCV} & ICCV 2021 & \ding{51} & \ding{55} & 90.03 & 67.52 & 74.37 & - \\
        LiSiam\textsuperscript{$\ast$} \cite{wang2022lisiam} & TIFS 2022 & \ding{55} & \ding{51}  & 90.36 & 72.59 & 82.06 & 76.52 \\
        AUNet \cite{Bai_2023_CVPR} & CVPR 2023 & \ding{51} & \ding{55} & 92.77 & 73.82 & 86.16 & 81.45 \\
        Delocate \cite{10.24963/ijcai.2024/648} & IJCAI 2024 & \ding{51} & \ding{55} & 91.30 & - & 84.00 & - \\
		KFD \cite{yu2025unlocking} & ICML 2025 & \ding{51} & \ding{55} & 94.71 & 79.12 & \underline{91.81} & - \\
		 \midrule
		DiffusionFF (Ours) & - & \ding{55} & \ding{51}  & \textbf{97.24} & \textbf{85.05} & \textbf{92.56} & \textbf{88.56} \\      
		\bottomrule
	\end{tabular}
\end{table*}

\subsubsection{Training Strategy} 
Given the significant differences between DSSIM map estimation and binary classification, jointly optimizing them often leads to suboptimal performance for both. To address this challenge, we propose a two-stage training strategy that fully decouples the two tasks at the training level, enabling dedicated and specialized optimization for each. In the first stage, we freeze the pretrained forgery detector and focus on training the conditioning projectors and the denoising diffusion model. This stage is guided by the loss function defined in Equation~\ref{eq3}, ensuring the model is effectively optimized for DSSIM map estimation. In the second stage, we freeze the forgery detector, conditioning projectors, and diffusion model, and instead optimize the artifact feature extractor alongside the classifier. This stage employs the standard Cross-Entropy (CE) loss to enhance classification performance. Our two-stage training strategy fully leverages the capabilities of each network component, ultimately maximizing the effectiveness of the DiffusionFF framework.
\section{Experiments}
\subsection{Setup}
\subsubsection{Datasets} 
We perform model training and intra-dataset evaluation on the widely used FaceForensics++ (FF++) dataset \cite{Rossler_2019_ICCV}, which contains 1,000 real facial videos and 4,000 fake ones generated by four manipulation techniques: DeepFakes (DF), Face2Face (F2F) \cite{Thies_2016_CVPR}, FaceSwap (FS), and NeuralTextures (NT) \cite{thies2019deferred}. For cross-dataset evaluation, we select four popular benchmarks: Celeb-DeepFake-v2 (CDF2) \cite{Li_2020_CVPR}, which applies advanced deepfake generation methods to YouTube celebrity videos; the DeepFake Detection Challenge (DFDC) \cite{dolhansky2020deepfake} and its Preview version (DFDCP) \cite{dolhansky2019dee}, which contain videos with various perturbations such as compression, downsampling, and noise; and FFIW-10K (FFIW) \cite{Zhou_2021_CVPR}, which incorporates multi-person scenarios.

\subsubsection{Baselines} 
We compare our approach with sixteen SOTA baselines for face forgery detection, including ten detection-only methods: DCL \cite{sun2022dual}, SBI \cite{Shiohara_2022_CVPR}, F\textsuperscript{2}Trans \cite{10004978}, SeeABLE \cite{Larue_2023_ICCV}, LAA-Net \cite{Nguyen_2024_CVPR}, RAE \cite{10.1007/978-3-031-72943-0_23}, FreqBlender \cite{zhou2024freqblender}, UDD \cite{fu2025exploring}, DFD-FCG \cite{Han_2025_CVPR}, and Effort \cite{yan2024orthogonal}; and six joint detection and localization methods: LRL \cite{chen2021local}, PCL+I2G \cite{Zhao_2021_ICCV}, LiSiam \cite{wang2022lisiam}, AUNet \cite{Bai_2023_CVPR}, Delocate \cite{10.24963/ijcai.2024/648}, and KFD \cite{yu2025unlocking}. These baselines cover diverse technical paradigms, from traditional convolutions to SOTA Large Language Models (LLMs). 

\subsubsection{Evaluation Metrics}
We evaluate detection performance using the Area Under the Receiver Operating Characteristic Curve (AUC), a widely adopted metric for face forgery detection. Video-level results are reported by averaging frame-level predictions across each video. To assess the quality of the estimated DSSIM maps, we employ four standard metrics for evaluating image reconstruction and generation: Peak Signal-to-Noise Ratio (PSNR), Structural Similarity (SSIM) \cite{ssim}, Learned Perceptual Image Patch Similarity (LPIPS) \cite{Zhang_2018_CVPR}, and Fréchet Inception Distance (FID) \cite{heusel2017gans}.

\begin{table}[t]	
	\centering\renewcommand\arraystretch{1.2}\setlength{\tabcolsep}{2.8pt}
	\belowrulesep=0pt\aboverulesep=0pt
    \caption{Cross-dataset DSSIM map estimation performance. Note that the DFDC and DFDCP datasets lack GT DSSIM maps because their real and fake video pairs are not spatially aligned. \label{cross-dataset-loc}}
	\begin{tabular}{c|c|cccc}
		\toprule
		\multirow{2}{*}{Dataset} & \multirow{2}{*}{Method} & 
		\multicolumn{4}{c}{Evaluation Metrics} \\
		\cmidrule(lr){3-6}
		& &\multicolumn{1}{c}{PSNR$\uparrow$} 
		&\multicolumn{1}{c}{SSIM$\uparrow$} 
		&\multicolumn{1}{c}{LPIPS$\downarrow$}
		&\multicolumn{1}{c}{FID$\downarrow$}\\
		\midrule
        \multirow{3}{*}{CDF2} & LiSiam \cite{wang2022lisiam} & 21.985 & 0.367 & 0.464 & 256.201 \\
        & LRL \cite{chen2021local} & 20.801 & 0.422 & 0.455 & 258.674 \\
		& DiffusionFF & \textbf{30.697} & \textbf{0.546} & \textbf{0.376} & \textbf{98.982} \\    
        \midrule
        \multirow{3}{*}{FFIW} & LiSiam \cite{wang2022lisiam} & 23.689 & 0.421 & 0.446 & 284.271 \\
        & LRL \cite{chen2021local} & 23.999 & 0.460 & 0.409 & 248.781 \\
		& DiffusionFF & \textbf{32.828} & \textbf{0.529} & \textbf{0.394} & \textbf{191.798} \\  
		\bottomrule
	\end{tabular}
\end{table}

\begin{figure}[t]
	\begin{center}
		\begin{minipage}{1\linewidth}
			{\includegraphics[width=1\linewidth]{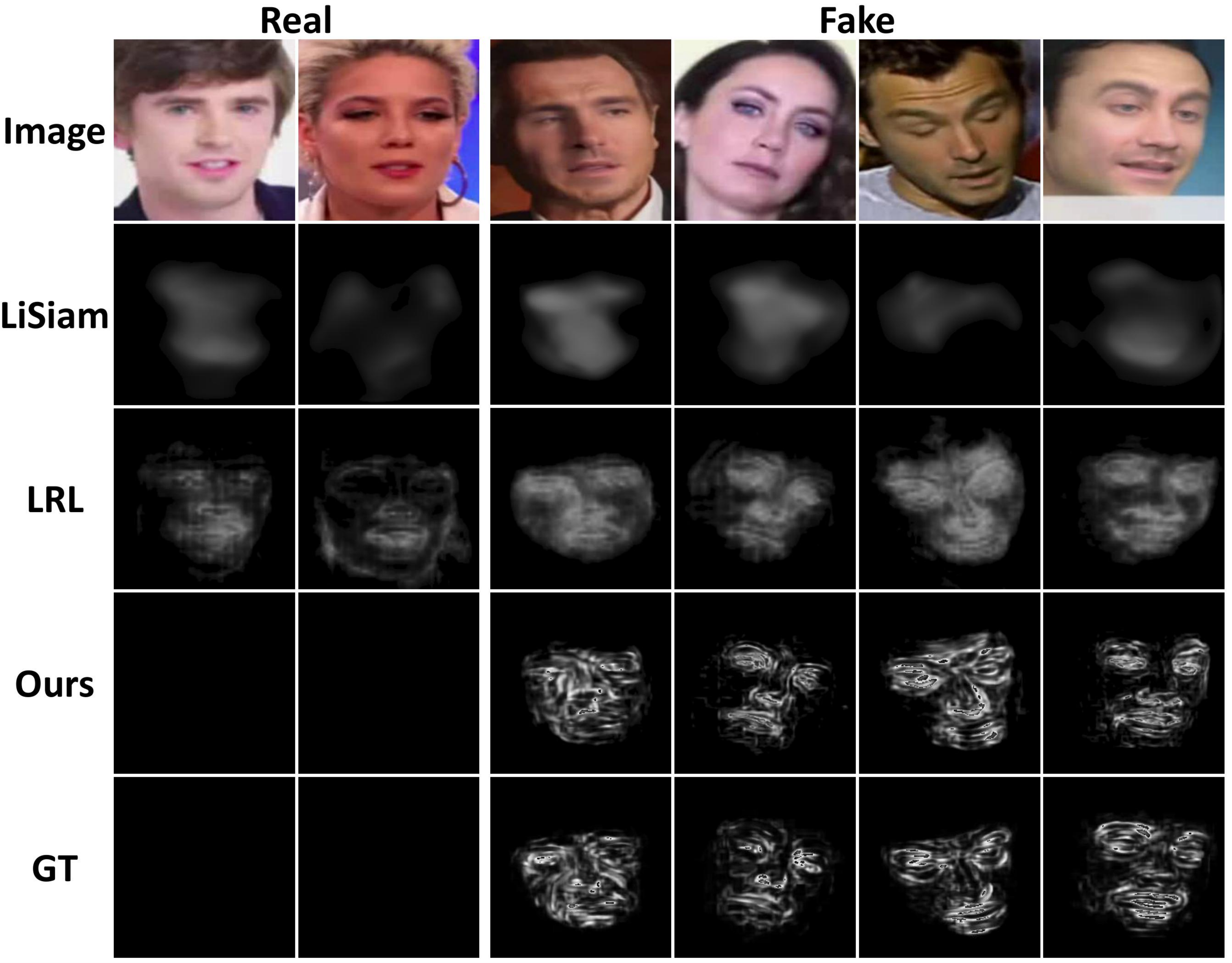}}
			\centering
		\end{minipage}
	\end{center}
	\caption{Qualitative DSSIM map estimation results on CDF2. DiffusionFF demonstrates strong generalization ability by producing more precise and fine-grained artifact localization results. \label{cdf_ssim}}
\end{figure}

\subsubsection{Implementation Details}
For forgery detection, we adopt the preprocessing techniques, data augmentation strategies, and testing pipeline developed by SBI \cite{Shiohara_2022_CVPR}. For artifact localization, we produce GT DSSIM maps on the FF++, CDF2, and FFIW datasets using the procedure outlined in Section~\ref{gt_ssim}. We employ a ConvNeXt-B network \cite{Liu_2022_CVPR} pretrained on the FF++ dataset as the forgery detector, which consists of four hierarchical stages. For the denoising diffusion model, we use a U-Net architecture \cite{ronneberger2015u} enhanced with a timestep encoder. Our two-stage training strategy is configured as follows. In the first stage, we train the conditioning projectors and the diffusion model for 100 epochs using the AdamW optimizer \cite{loshchilov2018decoupled}, with a batch size of 96 and an initial learning rate of $1 \times 10^{-4}$, which follows a cosine decay schedule. The total number of diffusion steps is set to $T = 50$. In the second stage, the artifact feature extractor and the classifier are trained for 5 epochs using AdamW, with a batch size of 128 and a fixed learning rate of $5 \times 10^{-5}$. All experiments are implemented in the PyTorch framework and executed on a cluster of eight NVIDIA RTX 3090 GPUs. For additional details, please kindly refer to the supplementary material.

\begin{table}[t]	
	\centering\renewcommand\arraystretch{1.2}\setlength{\tabcolsep}{5.2pt}
	\belowrulesep=0pt\aboverulesep=0pt
    \caption{Intra-dataset face forgery detection performance on the FF++ dataset. The proposed DiffusionFF framework yields the best overall results compared to existing SOTA approaches. \label{intra-dataset}}
	\begin{tabular}{c|cccc|c}
		\toprule
		\multirow{2}{*}{Method} & 
		\multicolumn{5}{c}{Test Set AUC (\%) $\uparrow$} \\
		\cmidrule(lr){2-6}
		&\multicolumn{1}{c}{DF} 
		&\multicolumn{1}{c}{F2F} 
		&\multicolumn{1}{c}{FS}
		&\multicolumn{1}{c|}{NT}
		&\multicolumn{1}{c}{FF++} \\
		\midrule
        LRL \cite{chen2021local} & \textbf{100} & \textbf{100} & 99.95 & 99.59 & 99.88 \\
		PCL+I2G \cite{Zhao_2021_ICCV} & \textbf{100} & 99.57 & \textbf{100} & 99.58 & 99.79 \\
        LiSiam \cite{wang2022lisiam} & \textbf{100} & 99.99 & 99.99 & 99.46 & 99.86 \\
		AUNet \cite{Bai_2023_CVPR} & \textbf{100} & 99.86 & 99.98 & 99.71 & 99.89 \\
		RAE \cite{10.1007/978-3-031-72943-0_23} & 99.60 & 99.10 & 99.20 & 97.60 & 98.90 \\
		\midrule
		DiffusionFF & \textbf{100} & \textbf{100} & 99.90 & \textbf{99.86} & \textbf{99.94}\\      
		\bottomrule
	\end{tabular}
\end{table}

\begin{figure*}[t]
	\begin{center}
		\begin{minipage}{1\linewidth}
			{\includegraphics[width=1\linewidth]{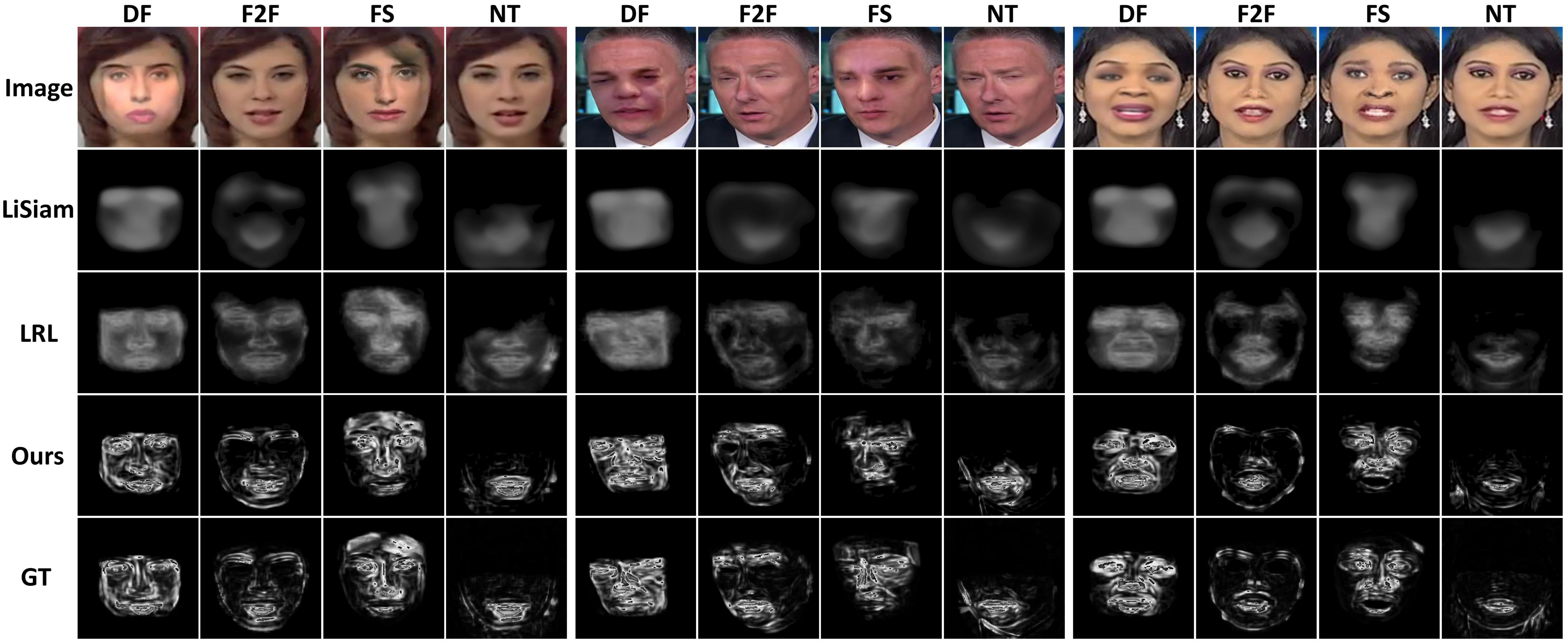}}
			\centering
		\end{minipage}
	\end{center}
	\caption{Qualitative DSSIM map estimation results on the FF++ dataset. Our method achieves the most visually faithful outcomes. \label{ff_ssim}}
\end{figure*}

\begin{table}[t]	
	\centering\renewcommand\arraystretch{1.2}\setlength{\tabcolsep}{6.2pt}
	\belowrulesep=0pt\aboverulesep=0pt
    \caption{Intra-dataset DSSIM map estimation performance on the FF++ dataset. Our method demonstrates superior performance compared to existing DSSIM-based localization approaches, with particularly significant improvement observed in the FID metric. \label{intra-dataset-loc}}
	\begin{tabular}{c|cccc}
		\toprule
		\multirow{2}{*}{Method} & 
		\multicolumn{4}{c}{Evaluation Metrics} \\
		\cmidrule(lr){2-5}
		&\multicolumn{1}{c}{PSNR$\uparrow$} 
		&\multicolumn{1}{c}{SSIM$\uparrow$} 
		&\multicolumn{1}{c}{LPIPS$\downarrow$}
		&\multicolumn{1}{c}{FID$\downarrow$}\\
		\midrule
        LiSiam \cite{wang2022lisiam} & 20.753 & 0.408 & 0.426 & 292.149 \\
        LRL \cite{chen2021local} & 23.608 & 0.595 & 0.326 & 241.625 \\
		DiffusionFF & \textbf{26.376} & \textbf{0.718} & \textbf{0.198} & \textbf{43.093} \\      
		\bottomrule
	\end{tabular}
\end{table}

\subsection{Performance}
\subsubsection{Cross-Dataset Evaluation}
Cross-dataset evaluation is a crucial protocol for assessing the generalization ability of face forgery detection models on previously unseen data. As shown in Table~\ref{cross-data}, we compare our method against recent SOTA approaches for face forgery detection. The proposed DiffusionFF framework outperforms all competing models, achieving the highest AUC scores across all unseen datasets, demonstrating its exceptional generalization capability. For artifact localization, we conduct both quantitative and qualitative comparisons with existing DSSIM-based techniques. The results in Table~\ref{cross-dataset-loc} and Figure~\ref{cdf_ssim} demonstrate the superior effectiveness of DiffusionFF in producing high-quality localization maps. This remarkable performance can be attributed to the novel application of diffusion models to estimate DSSIM maps, which enables the precise localization of fine-grained forgery clues and ultimately enhances detection capability.

\subsubsection{Intra-Dataset Evaluation}
In Table~\ref{intra-dataset}, we compare the intra-dataset detection performance of DiffusionFF with that of existing approaches, highlighting the superior discriminative power of our method. Furthermore, the quantitative results in Table~\ref{intra-dataset-loc} and qualitative outcomes in Figure~\ref{ff_ssim} validate the efficacy of DiffusionFF in synthesizing high-quality localization maps.

\begin{table}[t]	
	\centering\renewcommand\arraystretch{1.2}\setlength{\tabcolsep}{2.1pt}
	\belowrulesep=0pt\aboverulesep=0pt
    \caption{Ablation study on the diffusion model. Our method produces the highest-quality DSSIM maps, validating the effectiveness of our design choices. ``Cond.'' is short for conditioning. \label{abl_diffusion}}
	\begin{tabular}{c|cccc}
		\toprule
		\multirow{2}{*}{Method} & 
		\multicolumn{4}{c}{Evaluation Metrics} \\
		\cmidrule(lr){2-5}
		&\multicolumn{1}{c}{PSNR$\uparrow$} 
		&\multicolumn{1}{c}{SSIM$\uparrow$} 
		&\multicolumn{1}{c}{LPIPS$\downarrow$}
		&\multicolumn{1}{c}{FID$\downarrow$} \\
		\midrule
		Direct Regression & 24.341 & 0.584 & 0.252 & 172.213 \\
		Latent-Space Diffusion & 24.228 & 0.680 & 0.217 & 56.384 \\
		Final-Stage Cond. & 26.154 & 0.711 & 0.204 & 43.812 \\
            Decoder Cond. & 26.253 & 0.716 & 0.202 & 46.329 \\
		\midrule
		The Proposed & \textbf{26.376} & \textbf{0.718} & \textbf{0.198} & \textbf{43.093}\\ 
		\bottomrule
	\end{tabular}
\end{table}

\subsection{Model Analysis}
In this section, we present a comprehensive analysis of the DiffusionFF framework, including ablation studies on the diffusion model and fusion module, as well as evaluations of model generality and robustness. Additional ablation experiments, covering training and inference settings of the diffusion model, denoising seeds, and the two-stage training pipeline, are all detailed in the supplementary material.

\begin{table}[t]	
	\centering\renewcommand\arraystretch{1.2}\setlength{\tabcolsep}{2.4pt}
	\belowrulesep=0pt\aboverulesep=0pt
    \caption{Ablation study on the fusion module. The selected gating mechanism surpasses alternative fusion strategies in face forgery detection, confirming the soundness of our design choice. \label{abl_fusion}}
	\begin{tabular}{c|cccc|c}
		\toprule
		\multirow{2}{*}{Method} & 
		\multicolumn{4}{c|}{Test Set AUC (\%) $\uparrow$} &
            \multirow{2}{*}{Avg.} \\
		\cmidrule(lr){2-5}
		&\multicolumn{1}{c}{CDF2} 
		&\multicolumn{1}{c}{DFDC} 
		&\multicolumn{1}{c}{DFDCP}
		&\multicolumn{1}{c|}{FFIW} \\
		\midrule
		Addition & 95.79 & 84.68 & 91.56 & 88.38 & 90.10 \\
		Hadamard Product & 96.46 & 84.53 & 91.37 & 88.40 & 90.19 \\
		Concatenation & 96.66 & 85.00 & 91.87 & \textbf{89.01} & 90.64 \\
            Cross-Attention & 97.19 & 84.91 & 92.26 & 88.46 & 90.71 \\
		\midrule
		Gating Mechanism & \textbf{97.24} & \textbf{85.05} & \textbf{92.56} & 88.56 & \textbf{90.85}\\ 
		\bottomrule
	\end{tabular}
\end{table}

\subsubsection{Ablation Study on the Diffusion Model} 
We conduct a series of ablation experiments to evaluate the key design choices in our denoising diffusion model for fine-grained artifact localization. Specifically, our experiments investigate: (1) the effectiveness of iterative generation compared to direct regression; (2) the impact of utilizing diffusion models at the pixel level rather than in the latent space \cite{rombach2022high}; (3) the advantages of multi-stage conditioning over conditioning applied only at the final stage; and (4) the benefits of encoder-based conditioning relative to conditioning on the U-Net decoder. The quantitative DSSIM map estimation outcomes, as summarized in Table~\ref{abl_diffusion}, substantiate the validity and soundness of our architectural choices.

\subsubsection{Ablation Study on the Fusion Module} 
To assess the effectiveness of the gating mechanism for feature fusion, we perform comparative experiments against several alternative strategies, including addition, Hadamard product, concatenation, and cross-attention \cite{vaswani2017attention}. As shown in Table~\ref{abl_fusion}, the cross-dataset detection results demonstrate the superior ability of the gating mechanism to seamlessly integrate artifact-aware and high-level semantic features.

\begin{table}[t]	
	\centering\renewcommand\arraystretch{1.2}\setlength{\tabcolsep}{3pt}
	\belowrulesep=0pt\aboverulesep=0pt
    \caption{Evaluation of model generality. Integrating DiffusionFF as an auxiliary module consistently enhances the performance of existing forgery detectors, demonstrating its effectiveness as a plug-and-play improvement for face forgery detection models. \label{others}}
	\begin{tabular}{c|c|cccc}
		\toprule
		\multirow{2}{*}{Method} & 
		\multirow{2}{*}{Params} & 
		\multicolumn{4}{c}{Test Set AUC (\%) $\uparrow$} \\
		\cmidrule(lr){3-6}
		&\multicolumn{1}{c|}{} 
		&\multicolumn{1}{c}{CDF2} 
		&\multicolumn{1}{c}{DFDC} 
		&\multicolumn{1}{c}{DFDCP}
		&\multicolumn{1}{c}{FFIW} \\
		\midrule
            EfficientNet-B4 & 19M & 93.18 & 72.42 & 86.15 & 84.83 \\
		+ DiffusionFF & 32M & 95.54 & 82.19 & 89.03 & 86.37 \\
            \midrule
            Swin-B & 88M & 95.59 & 82.81 & 89.74 & 85.38 \\
		+ DiffusionFF & 101M & 96.93 & 84.54 & 90.78 & 87.01 \\
		\midrule
		ConvNeXt-B & 89M & 96.08 & 84.26 & 90.20 & 87.67 \\
		+ DiffusionFF & 102M & \textbf{97.24} & \textbf{85.05} & \textbf{92.56} & \textbf{88.56} \\
		\bottomrule
	\end{tabular}
\end{table}

\begin{figure}[t]
	\begin{center}
		\begin{minipage}{1\linewidth}
			{\includegraphics[width=1\linewidth]{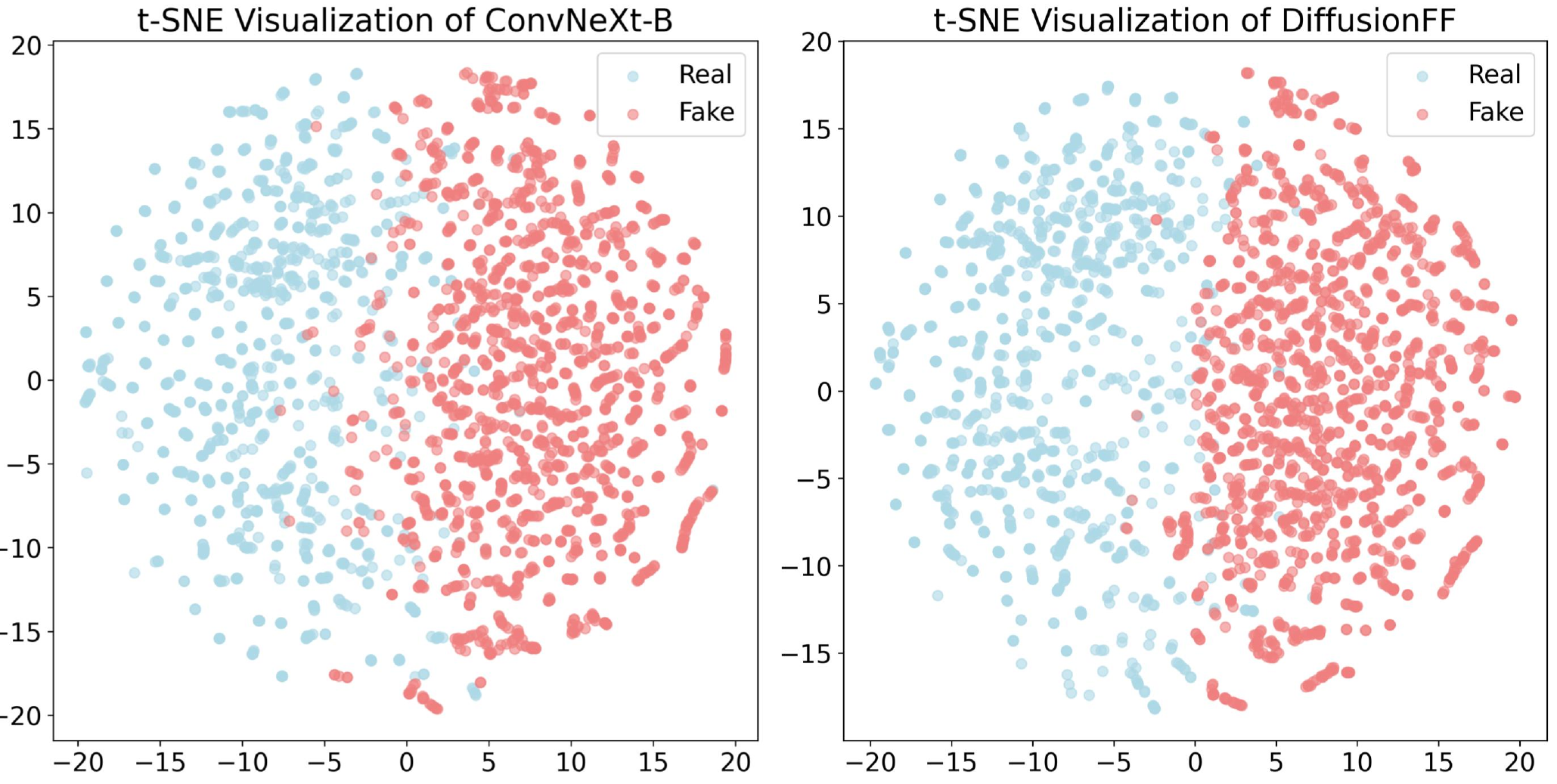}}
			\centering
		\end{minipage}
	\end{center}
	\caption{t-SNE visualizations of feature representations extracted by ConvNeXt and DiffusionFF on the CDF2 dataset. Compared to ConvNeXt, DiffusionFF yields more distinct and well-separated clusters, demonstrating its superior discriminative capability. \label{tsne}}
\end{figure}

\subsubsection{Evaluation of Model Generality} 
We evaluate the generality of the DiffusionFF framework by examining its ability to enhance the detection performance of various forgery detectors. Our study considers two CNN-based models, EfficientNet \cite{tan2019efficientnet} and ConvNeXt \cite{Liu_2022_CVPR}, as well as the Transformer-based Swin model \cite{Liu_2021_ICCV}. For each detector, we compare its baseline performance against its performance when augmented with DiffusionFF, which provides additional artifact localization maps serving as visual explanations. The Vision Transformer (ViT) \cite{dosovitskiy2020vit} is excluded because its single-scale feature representation is incompatible with DiffusionFF's multi-scale conditioning mechanism. As shown in Table~\ref{others}, all detectors exhibit consistent and substantial gains in cross-dataset detection performance when combined with DiffusionFF. Furthermore, t-SNE visualizations \cite{van2008visualizing} in Figure~\ref{tsne} illustrate that DiffusionFF enables a clearer separation between real and fake samples than using the detector alone. These results confirm that integrating DiffusionFF as an auxiliary module significantly improves the performance of existing detectors, demonstrating its effectiveness as an explainable and plug-and-play enhancement for forgery detection models.

\begin{figure}[t]
	\begin{center}
		\begin{minipage}[t]{1\linewidth}
			\begin{minipage}[t]{0.49\linewidth}
				{\includegraphics[width=1\linewidth]{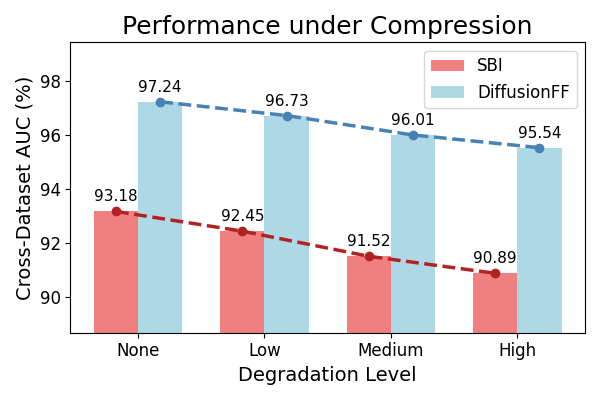}}
				{\includegraphics[width=1\linewidth]{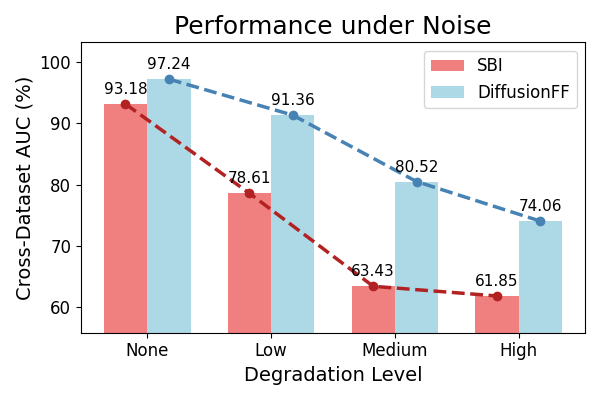}}
				\centering
				
			\end{minipage}
			\begin{minipage}[t]{0.49\linewidth}
				{\includegraphics[width=1\linewidth]{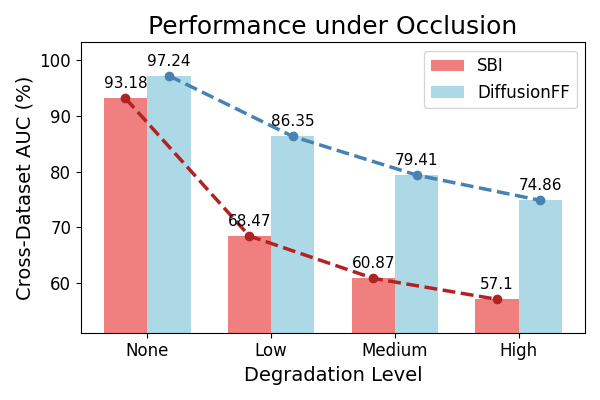}}
				{\includegraphics[width=1\linewidth]{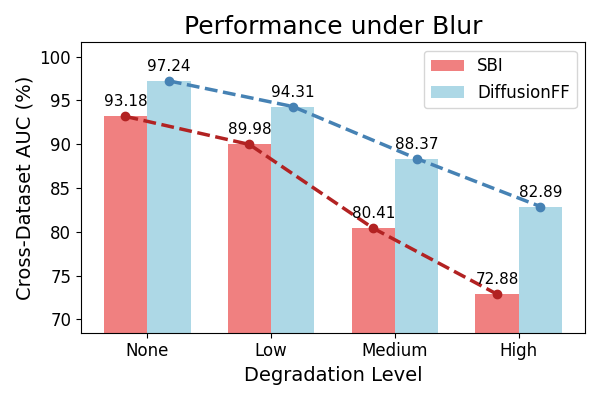}}
				\centering
				
			\end{minipage}
		\end{minipage}
	\end{center}
	\caption{Evaluation of model robustness under various degradation conditions. DiffusionFF shows smaller performance declines than SBI \cite{Shiohara_2022_CVPR} across all degradation types and severity levels. \label{degradation}}
\end{figure}

\subsection{Evaluation of Model Robustness}
We evaluate our model's robustness across four real-world degradations, each applied at three severity levels: JPEG compression (quality factors of 95, 85, and 75), random occlusion (masking 2\%, 5\%, and 10\% of pixels), Gaussian noise (standard deviations of 0.01, 0.05, and 0.1), and Gaussian blur (kernel/sigma values of 5/1, 7/2, and 9/3). As shown in Figure~\ref{degradation}, the cross-dataset assessment results on the CDF2 dataset demonstrate that DiffusionFF consistently outperforms SBI \cite{Shiohara_2022_CVPR} across all degradation conditions.

\section{Conclusion}
We propose DiffusionFF, a novel framework that simultaneously addresses the dual challenges of face forgery detection and fine-grained artifact localization. DiffusionFF introduces an innovative use of denoising diffusion models to estimate detailed DSSIM maps, enabling precise localization of fine-grained forgery clues and substantially improving detection capability. Extensive experiments demonstrate that our method not only achieves SOTA detection performance but also surpasses existing artifact localization approaches in revealing manipulation traces, underscoring its superior effectiveness, reliability, and explainability.

{
    \small
    \bibliographystyle{ieeenat_fullname}
    \bibliography{main}
}

\clearpage
\setcounter{page}{1}
\maketitlesupplementary

\section{Additional Implementation Details}
\subsection{SBI}
The SBI framework \cite{Shiohara_2022_CVPR} synthesizes pseudo-fake facial images from real samples by simulating the deepfake generation process. It consists of two main components: the Source-Target Generator (STG) and the Mask Generator (MG). The STG applies a series of image transformations to create a pseudo-source and a pseudo-target image, while the MG produces a blending mask based on pre-detected facial landmarks, with augmentation to enhance diversity. The pseudo-source and pseudo-target images are then combined using the blending mask to generate a pseudo-fake facial image. In this study, we employ the SBI framework to augment the training data by integrating SBI-generated samples with those from the original FF++ dataset \cite{Rossler_2019_ICCV}. A key advantage of this approach is that the pseudo-fake images are perfectly aligned with their real counterparts, allowing for the precise computation of GT DSSIM maps.

\subsection{Preprocessing}
During training, we extract 32 frames from each real video and 8 frames from each fake video to maintain a balanced ratio between positive and negative samples, as each real video corresponds to four manipulated ones. Additionally, we sample 8 frames from each real video to generate pseudo-fake samples using the SBI framework, which are incorporated into the training set to enhance data diversity. During testing, we uniformly sample 32 frames per video. Facial landmarks are extracted using Dlib's 81-point predictor \cite{king2009dlib}, which is utilized only during the training phase. For face detection, RetinaFace \cite{Deng_2020_CVPR} is employed to obtain facial bounding boxes. During training, each detected face is cropped with a random margin ranging from 4\% to 20\%, while a fixed margin of 12.5\% is applied during inference.

\subsection{Data Augmentation}
The image processing toolbox introduced in \cite{buslaev2020albumentations} is employed for data augmentation. Within the STG module of the SBI framework, transformations including RGBShift, HueSaturationValue, RandomBrightnessContrast, Downscale, and Sharpen are applied to generate pseudo source and target images. During training, all samples are augmented using operations such as ImageCompression, RGBShift, HueSaturationValue, and RandomBrightnessContrast. These augmentations expose the model to a broader range of visual variations, thereby enhancing its generalization capability.

\subsection{Additional Details}
The ConvNeXt-B detector \cite{Liu_2022_CVPR} is trained on the FF++ dataset with SBI-based data augmentation. Training is conducted for 200 epochs using the AdamW optimizer \cite{loshchilov2018decoupled}, with a batch size of 64 and an initial learning rate of $5\times 10^{-5}$. To promote stable convergence, a linear learning rate decay is applied starting from epoch 100. For the diffusion model, we employ a standard seven-stage U-Net architecture \cite{ronneberger2015u} with channel dimensions configured as [64, 64, 128, 256, 128, 64, 64]. For the noise schedule, the per-step variances $\{\beta_t\}_{t=1}^{T}$ increase linearly from 0.02 to 0.4. The GT DSSIM maps are generated using a local square window of size $7\times 7$. For frames containing multiple detected faces, the classification model is applied to each face, and the maximum score is taken as the frame-level confidence.

\begin{figure}[t]
	\begin{center}
		\begin{minipage}{1\linewidth}
			{\includegraphics[width=1\linewidth]{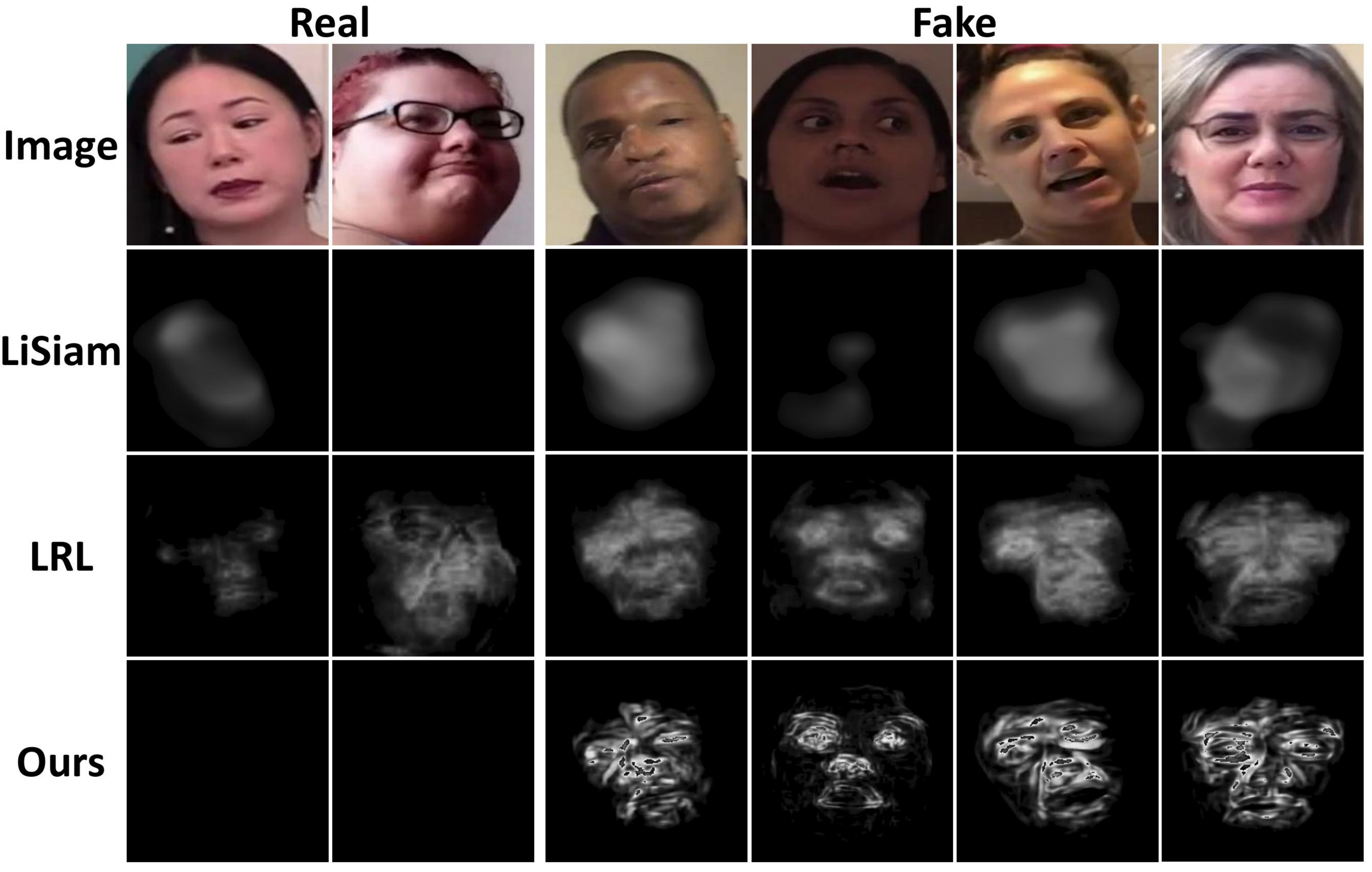}}
			\centering
		\end{minipage}
	\end{center}
	\caption{Qualitative DSSIM map estimation results on the DFDC dataset. Our DiffusionFF framework generates more fine-grained and detailed artifact localization maps than competing approaches. Note that GT DSSIM maps are unavailable for this dataset. \label{dfdc}}
\end{figure}

\section{Additional Experiments}
\subsection{Additional Qualitative Evaluation}
In the DFDC \cite{dolhansky2020deepfake} and DFDCP \cite{dolhansky2019dee} datasets, the real and fake video pairs are not spatially aligned, which prevents the generation of GT DSSIM maps for quantitative evaluation. To assess the generalization capability of our DiffusionFF framework under this situation, we provide a qualitative comparison against existing DSSIM-based approaches on the DFDC dataset, as shown in Figure~\ref{dfdc}. Although GT DSSIM maps are unavailable, the visual results clearly indicate that DiffusionFF produces the most fine-grained and detailed outcomes, underscoring its superior effectiveness in generating high-quality artifact localization maps.

\begin{figure}[t]
	\begin{center}
		\begin{minipage}[t]{1\linewidth}
			\begin{minipage}[t]{0.49\linewidth}
				{\includegraphics[width=1\linewidth]{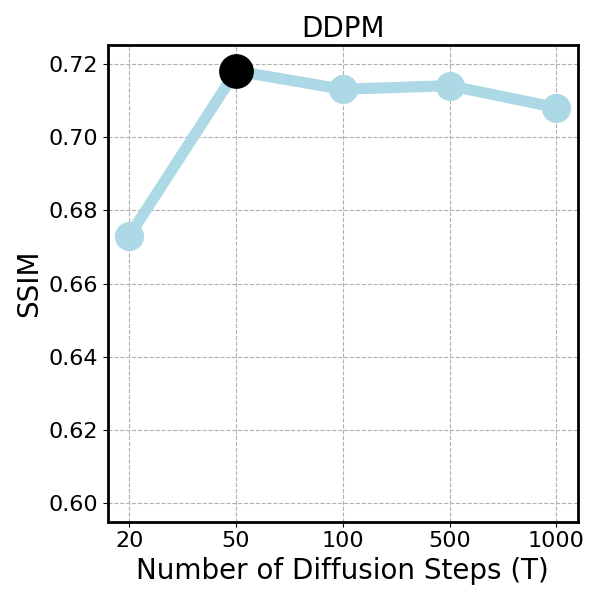}}
				\centering
				
			\end{minipage}
			\begin{minipage}[t]{0.49\linewidth}
				{\includegraphics[width=1\linewidth]{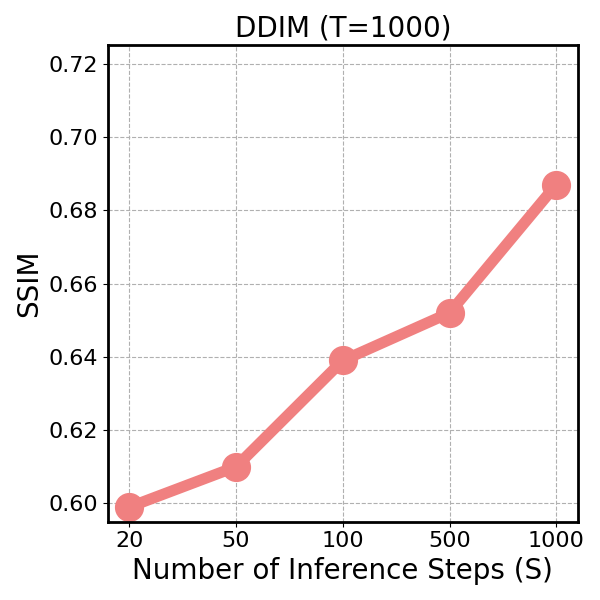}}
				\centering
				
			\end{minipage}
		\end{minipage}
	\end{center}
	\caption{Ablation study on the training and inference settings of the diffusion model. The DDPM framework with $T=50$ delivers the best performance, marked by the black dot in the left figure. \label{step}}
\end{figure}

\subsection{Additional Ablation Studies}
\subsubsection{Training \& Inference of the Diffusion Model}
In this study, we employ the standard DDPM framework \cite{ddpm}, where the number of inference steps is set equal to the number of training diffusion steps ($T$). We begin by examining how varying $T$ affects performance. The quantitative results shown on the left of Figure~\ref{step} indicate that our model achieves optimal performance at $T=50$. This outcome contrasts with the common expectation that larger $T$ values generally lead to better generation quality. We hypothesize that this discrepancy arises because our conditional DSSIM estimation task differs fundamentally from traditional RGB image generation, thereby shifting the optimal hyper-parameter landscape. To further validate our framework choice, we also evaluate the Denoising Diffusion Implicit Model (DDIM) framework \cite{ddim}, which enables faster inference by skipping diffusion steps. In this experiment, we use a model trained with $T=1000$ and perform inference using $S$ sampling steps. As shown on the right of Figure~\ref{step}, while DDIM performance improves with increasing $S$, it consistently underperforms the DDPM $T=1000$ baseline, let alone the superior DDPM $T=50$ setup. Collectively, these findings firmly validate our choice to use the DDPM framework with a small timestep of $T=50$.

\begin{figure}[t]
	\begin{center}
		\begin{minipage}{1\linewidth}
			{\includegraphics[width=1\linewidth]{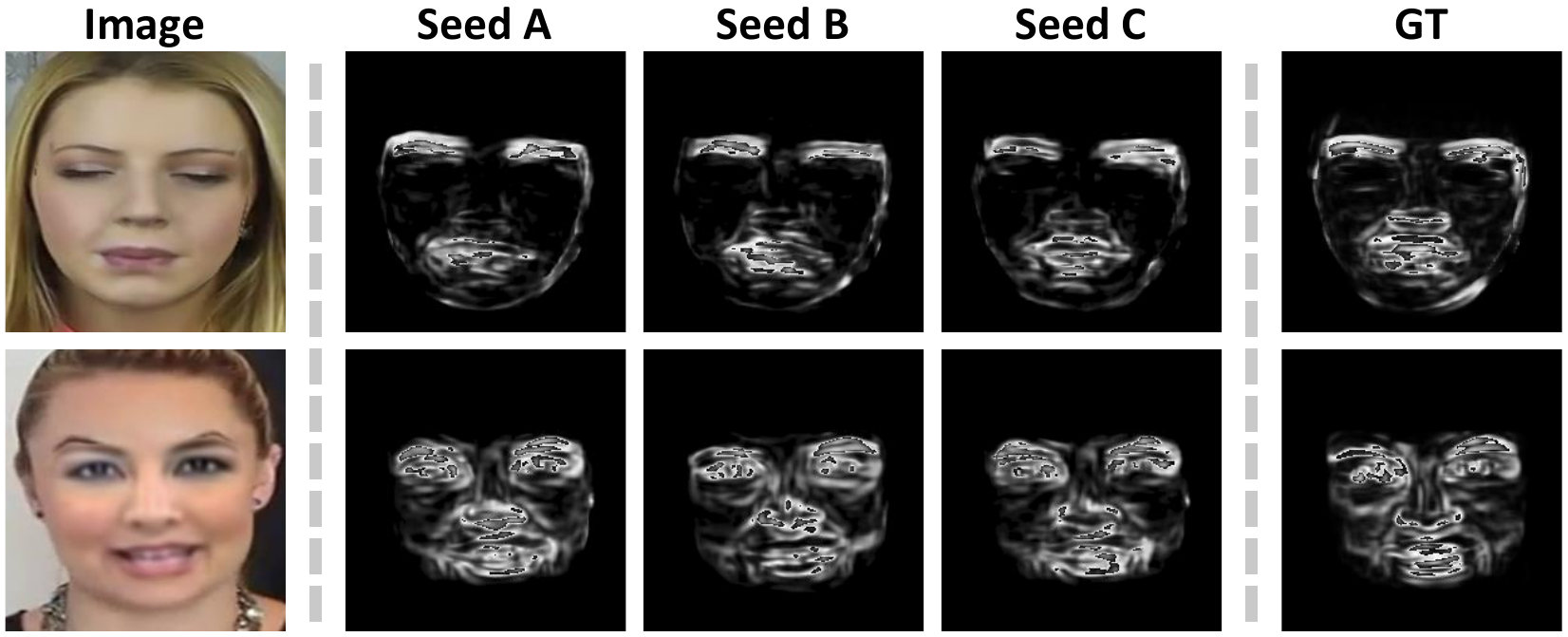}}
			\centering
		\end{minipage}
	\end{center}
	\caption{Ablation study on denoising seeds. Our method consistently generates stable localization maps across different seeds.\label{seed}}
\end{figure}

\subsubsection{Denoising Seeds}
We evaluate the impact of random seed variation on DSSIM map estimation performance. As illustrated in Figure~\ref{seed}, our DiffusionFF framework consistently generates stable and reliable artifact localization maps across different seeds.

\begin{table}[t]	
	\centering\renewcommand\arraystretch{1.2}\setlength{\tabcolsep}{6.8pt}
	\belowrulesep=0pt\aboverulesep=0pt
	\begin{tabular}{c|cccc}
		\toprule
		\multirow{2}{*}{Method} & 
		\multicolumn{4}{c}{Test Set AUC (\%) $\uparrow$} \\
		\cmidrule(lr){2-5}
		&\multicolumn{1}{c}{CDF2} 
		&\multicolumn{1}{c}{DFDC} 
		&\multicolumn{1}{c}{DFDCP}
		&\multicolumn{1}{c}{FFIW} \\
		\midrule
            Single-Stage & 95.47 & 83.54 & 90.27 & 87.69 \\
		Two-Stage & \textbf{97.24} & \textbf{85.05} &      \textbf{92.56} & \textbf{88.56} \\ 
		\bottomrule
	\end{tabular}
    \caption{Ablation study on the two-stage training strategy. \label{training}}
\end{table}

\subsubsection{Two-Stage Training Strategy}
We compare our two-stage training strategy against a single-stage baseline that jointly optimizes DSSIM map estimation and binary classification. As shown in Table~\ref{training}, our approach yields significantly better results, demonstrating the effectiveness of training these two tasks separately.

\section{Discussions}
In this section, we discuss three critical questions: (1) Can GT DSSIM maps serve as inputs to the artifact feature extractor during our second training stage? (2) Is the ControlNet framework \cite{zhang2023adding} capable of estimating DSSIM maps? and (3) What causes the joint training of the forgery detector and diffusion model from scratch to fail? Additionally, we summarize the strengths and limitations of our work.

\subsection{GT DSSIM Maps As Inputs During Training}
In Figure 2 of the main text, we demonstrate a clear positive correlation: the higher the quality of the DSSIM maps integrated into the detection network, the greater the resulting performance gains. This observation naturally leads to the question: what if GT DSSIM maps were used as inputs to the artifact feature extractor during our second training stage? To explore this, we conduct an experiment using GT DSSIM maps and obtain a 96.72\% AUC on the CDF2 dataset \cite{Li_2020_CVPR}, which is lower than the 97.24\% AUC achieved with diffusion-estimated maps. This performance drop is primarily due to a mismatch in value distributions: GT DSSIM maps (e.g., all zeros for real samples) differ significantly from those estimated by the diffusion model. As a result, a model trained with GT DSSIM maps struggles to generalize when faced with diffusion-estimated inputs during inference, ultimately leading to degraded performance.

\subsection{ControlNet for DSSIM Map Estimation}
ControlNet is a widely adopted framework for conditional image generation that integrates an auxiliary conditioning network to guide a pretrained, frozen Stable Diffusion model \cite{rombach2022high}. \textbf{In our study, we attempted to train a ControlNet to estimate DSSIM maps but encountered training collapse, where the network produces identical, content-irrelevant outputs regardless of input variation}. We attribute this failure to a fundamental domain mismatch: the pretrained Stable Diffusion model, optimized for natural RGB image synthesis, struggles to adapt to single-channel, grayscale DSSIM maps due to its strong prior knowledge. To overcome this limitation, we propose an inverted strategy that contrasts with the conventional ControlNet paradigm. Instead of training a new conditioning network, we employ a pretrained forgery detector as a fixed conditioning model, leveraging its capacity to capture forgery-related features. By freezing the conditioning network and training the diffusion model instead, we shift the learning focus to the generative model itself. This reversal allows the diffusion model to effectively utilize the detector's specialized features, enabling precise and stable estimation of DSSIM maps.

\subsection{Training Detector and Diffusion From Scratch}
As noted in Section 3.2.2, jointly training the forgery detector and diffusion model from scratch leads to training collapse. Here, we analyze this failure in detail. \textbf{During training, the model converges to a trivial solution: producing entirely black images regardless of input.} This collapse arises because the GT DSSIM maps for real images are uniformly black. Consequently, predicting black results in zero loss for real samples, allowing the model to minimize training loss without learning the more complex task of detecting and localizing artifacts in fake images. This behavior highlights that the DSSIM estimation task intrinsically depends on forgery-related features. Without explicit guidance from these features, the diffusion model fails to learn the distinction between real and fake faces. To address this issue, we employ a pretrained forgery detector to extract forgery-specific features that guide the diffusion model in generating fine-grained artifact localization maps.

\subsection{Strengths \& Limitations}
\subsubsection{Strengths}
The proposed DiffusionFF framework exhibits three key advantages. First, it enables precise localization of fine-grained forgery clues, thereby enhancing model explainability and fostering greater user trust. Second, by integrating the generated artifact localization maps into the detection pipeline, DiffusionFF significantly improves overall detection performance. Third, owing to its strong generality, DiffusionFF can serve as an explainable, plug-and-play enhancement for existing face forgery detection models.

\subsubsection{Limitations}
Despite its many advantages, the DiffusionFF framework also presents two limitations. First, the inference process of the diffusion model is time-consuming. Second, our method is not designed for entirely synthesized images.

\end{document}